\pdfoutput=1

\documentclass[11pt]{article}

\usepackage[final]{ACL2023}

\usepackage{times}
\usepackage{latexsym}

\usepackage[T1]{fontenc}

\usepackage[utf8]{inputenc}

\usepackage{microtype}

\usepackage{inconsolata}
\usepackage{enumitem}
\usepackage{calc}
\usepackage{tcolorbox}
\usepackage{graphicx}
\usepackage{booktabs}
\usepackage{amsmath}
\usepackage{amssymb}
\usepackage{amsthm}
\usepackage{makecell}
\usepackage{xcolor} %
\usepackage{placeins}
\usepackage{tcolorbox}
    \tcbuselibrary{breakable}
\usepackage{verbatim}

\title{Probing Semantic Routing in Large Mixture-of-Expert Models}

\author{
    Matthew Lyle Olson$^{1,\dagger}$ \:
    Neale Ratzlaff$^{1,\dagger}$ \:
    Musashi Hinck$^{1,\dagger}$ \\
    \textbf{Man Luo}$^1$ \:
    \textbf{Sungduk Yu}$^1$ \:
    \textbf{Chendi Xue}$^2$ \:
    \textbf{Vasudev Lal}$^1$ 
    \vspace{0.2cm} \\
    $^1$Intel Labs \: $^2$Intel Corporation \\
}

\begin{document}
\maketitle
\begin{abstract}
In the past year, large ($>100$B parameter) mixture-of-expert (MoE) models have become increasingly common in the open domain. While their advantages are often framed in terms of efficiency, prior work has also explored functional differentiation through routing behavior. We investigate whether expert routing in large MoE models is influenced by the \textit{semantics} of the inputs. To test this, we design two controlled experiments. First, we compare activations on sentence pairs with a shared target word used in the same or different senses. Second, we fix context and substitute the target word with semantically similar or dissimilar alternatives. Comparing expert overlap across these conditions reveals clear, statistically significant evidence of \textit{semantic routing} in large MoE models.
\end{abstract}

\begin{figure}[t]
    \centering
    \includegraphics[width=.88\linewidth]{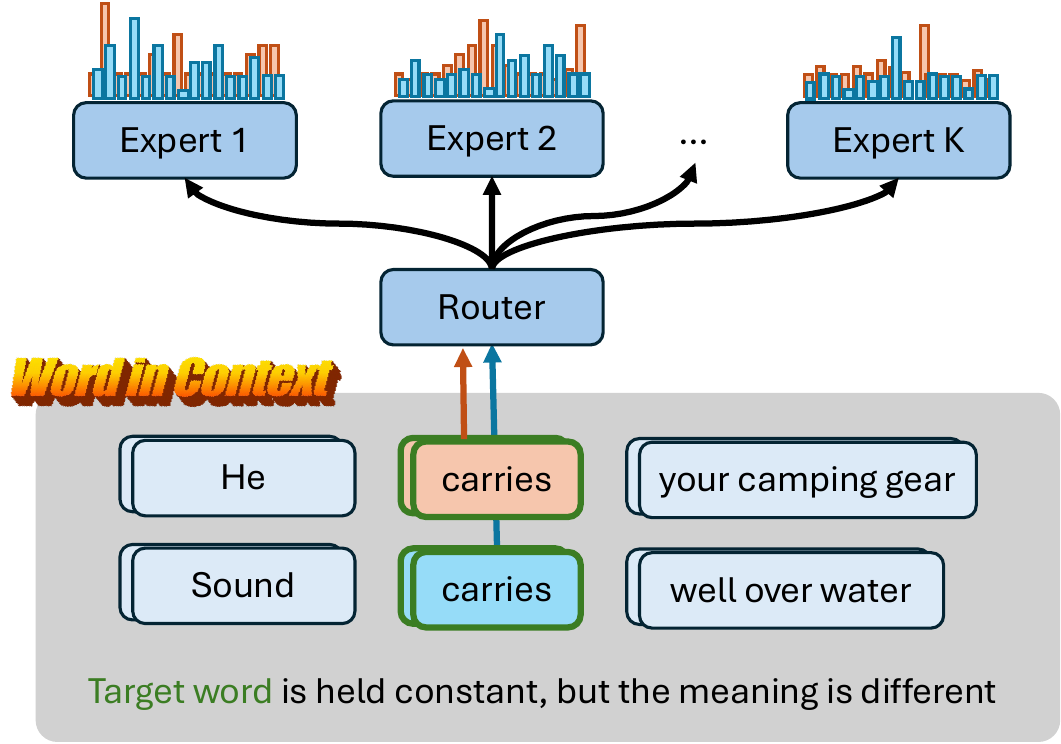} 
\par\vspace{.5em}
    \includegraphics[width=.88\linewidth]{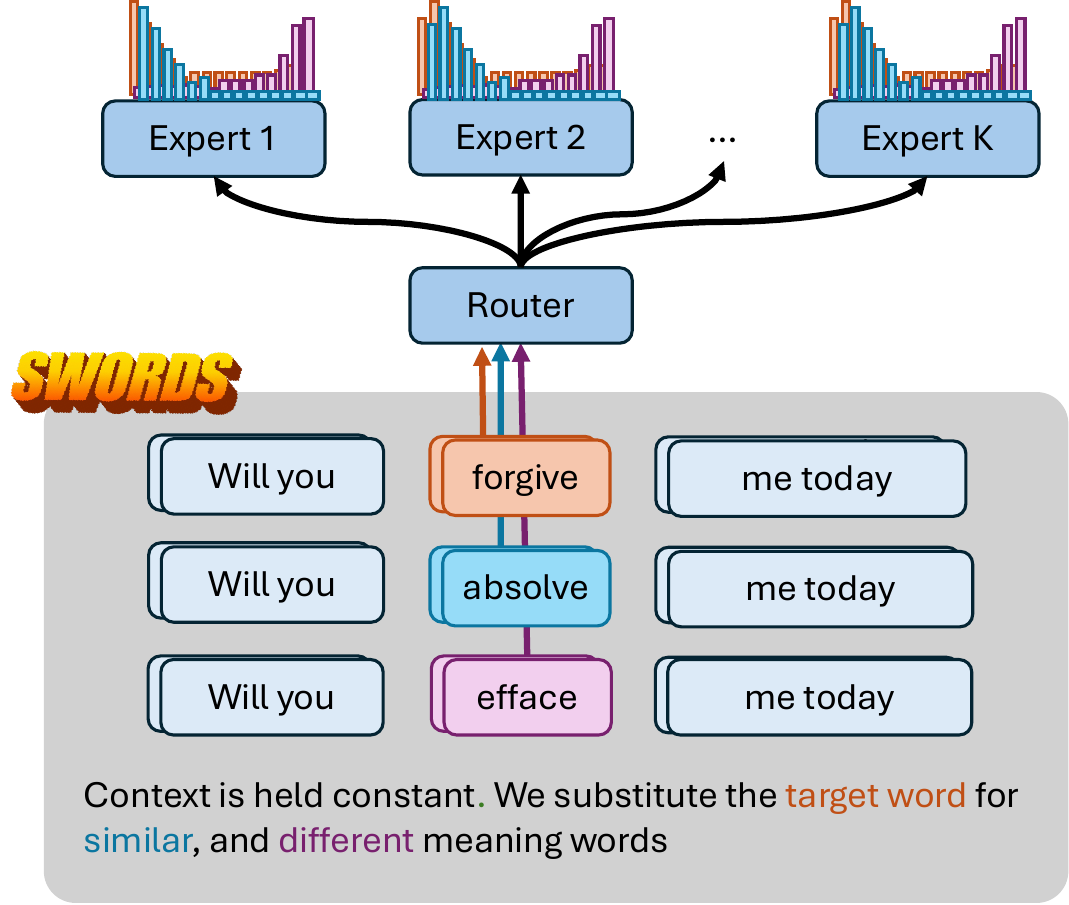} 
    \caption{\textbf{Summary of Experimental Design}. We compare expert routing patterns in two controlled experiments. \textit{Top:} we hold the \textcolor{green}{target word} constant, and change the context to either change the meaning of the \textcolor{green}{target word} or keep it the same. \textit{Bottom:} we hold context constant, and substitute the \textcolor{orange}{target word} for a \textcolor{blue}{similar-meaning} or \textcolor{pink}{different-meaning} word.}
    \label{fig:figure1}
    \vspace{-1em}
\end{figure}

\section{Introduction}

Since their popularization in \citet{fedus2022switch}, the Mixture-of-Experts (MoE) architecture \citep{ jacobs1991adaptive} has been integrated into many frontier large language models (LLMs) \citep{lieber2024jamba,jiang2024mixtral,liu2024deepseek,guo2025deepseek,meta2025llama4}. 
The MoE architecture offers the ability to train far larger models than would normally be possible with dense architectures. Designers can then modulate performance by varying the number of active experts to access a greater portion of the greater model, with the trend being to increase the total and active expert counts.

Several prior studies have explored expert activation patterns in MoE models, hypothesizing that each expert may specialize in specific domains, tasks, or topics~\cite{zoph2022stmoe,jiang2024mixtral,xue2024openmoe}. 
While it is intuitive to expect some degree of semantic specialization, previous research has not found clear evidence of routing on the basis of semantics, concluding instead that expert activation is primarily token-dependent rather than being driven by deeper semantic relationships.

Given that recent large-scale MoE models have achieved state-of-the-art performance while increasing total expert counts, 
we investigate whether these models' expert routing behavior exhibits semantic specialization.
We design two controlled experiments. First, we use a word sense disambiguation (WSD) task from the WiC benchmark~\cite{pilehvar2018wic}, where the same target word appears in two different sentences, either preserving or changing its meaning. This allows us to measure whether expert activation remains stable when the word's sense is preserved. Second, we study a complementary setting using the lexical substitution benchmark SWORDS~\cite{lee2021swords}, where we fix the surrounding context but vary the target word, comparing expert overlap between semantically similar and dissimilar word substitutions. We compare the rate of overlap, between models with differing numbers of active and total experts, via a normalized metric based on Cohen's $\kappa$ that controls for the baseline probability of overlap.

We apply these experiments to six MoE models from three model families: DeepSeek-R1\cite{guo2025deepseek}, DeepSeek-V2-Lite \cite{liu2024deepseek}, Mixtral-8x7B, Mixtral 8x22B \cite{jiang2024mixtral}, Llama-4-Scout and Llama-4-Maverick \cite{meta2025llama4}. For all models, we find that the rate of expert overlap is significantly higher when the meaning of the target word is equal in two sentences than when the meaning of the target word is different.
We also find that model scale influences the strength of this specialization: larger models generally exhibit stronger semantic routing signals-- with Llama-4 Scout~\cite{meta2025llama4} standing out as an exception, showing a pronounced effect despite its smaller total parameter count. 
Finally, semantic differentiation in expert routing is most prominent in the middle layers, where DeepSeek-R1 exhibits the clearest and most consistent specialization pattern.

In summary, our contributions are threefold:
(1) We design two complementary semantic probing setups, based on word sense disambiguation and semantic substitution, to systematically assess expert specialization in recent MoE models.
(2) We introduce an expert overlap metric to quantify routing similarity and demonstrate its alignment with lexical relationships.
(3) We conduct extensive experiments across three MoE model families (DeepSeek, Mixtral, and Llama-4) at various scales, uncovering clear empirical evidence of semantic routing and highlighting its dependence on model size and layer depth.

\section{Related Work}
Current research on expert specialization in MoE models is sparse, yet available studies reveal little evidence of semantic-level differentiation. For example, \citet{xue2024openmoe} tracked token routing patterns across datasets segmented by different topics, languages, and tasks, but failed to find any coherent pattern at such high-level semantics. Rather, they found indications of token-level specialization, mainly concerning low-level semantic features like special characters or auxiliary verbs. Similar findings have been reported in studies using independently developed MoE models \citep[e.g.,][]{zoph2022stmoe, jiang2024mixtral, fan2024towards}.

While some neuroscience research has provided evidence that the brain functions like a Mixture of Experts \citep{stocco2010conditional,o2021and}---suggesting the possibility of semantic-level specialization---other studies have shown that MoE models with random routing can perform comparably to those using the more common top-k routing approach \citep{roller2021hash, zuo2021taming, ren2023pangu}. One potential explanation for these mixed results is that prior models (using 8 to 32 experts) might not have been sufficiently expressive to capture fine-grained specialization patterns. The recently-released DeepSeek V3 and Llama 4 Maverick, featuring an extensive network of experts (256 and 128 routed specialists, respectively), provide us with a unique opportunity. Hence, in this study, we test whether a more capable MoE architecture exhibits semantic-level expert specialization.

\section{Experiment Settings} %
\subsection{Evaluation Datasets}
\paragraph{Words-in-Context} 

We leverage polysemy to test for semantic specialization in expert activation patterns. If words that are written the same but have different meanings are routed differently, then this is evidence that routing occurs based on meaning. To test this hypothesis, we use the WiC dataset~\cite{pilehvar2018wic} (CC BY-NC 4.0), which consists of two types of paired sentences: 1) pairs where a target word has the same sense and 2) pairs where the target word has different senses across sentences.
\paragraph{SWORDS}
We construct a complementary scenario to the WiC experiment, where we test the degree of expert overlap on semantically similar, lexically different input phrases. 
To do so, we leverage SWORDS~\cite{lee2021swords} (CC-BY-3.0-US) a lexical substitution benchmark where the corresponding dataset provides semantically annotated sentence pairs with single- and multi-token phrase replacements. We use the SWORDS dataset to construct triples of sentences where a target word is substituted either for a semantically equivalent word or a non-equivalent one. 
We show examples of both experimental settings in Figure~\ref{fig:figure1}, and an example of such a triplet with target words as follows: %
\begin{description}[leftmargin=!,labelwidth=\widthof{\textit{Equivalent}}]
    \item[\textit{Original}]: \small{"My last show was glorious!" Tasha said.}
    \item[\textit{Equivalent}]: \small{"My last show was splendid!" Tasha said.}
    \item[\textit{Different}]: \small{"My last show was notable!" Tasha said.}
\end{description}

For both datasets, we construct the following prompts. For each target words and sentence, we prompt the non-reasoning models with:
``\textit{<user> Please define \{target word\} in this context <assistant> Sure! Here is the definition of the word \{target word\}}"

Alternatively, for the reasoning models we use:
``\textit{<user> Please define \{target word\} in this context <assistant> <think> Okay, so I need to figure out the meaning of the word \{target word\}}'' to ensure the word in question is analyzed instead of additional thinking tokens.

\subsection{Models} 

We analyze three recent families of MoE-based models in our study, an overview of parameter and expert counts is provided in Table~\ref{tab:model_size}. \\
\textbf{DeepSeek} MoE models represent the largest and smallest models that we study. DeepSeek-R1 has the highest parameter count (671B) and number of active experts (8/256), while DeepSeek-v2-Lite has just 15.7B parameters and 8/64 active experts.\\
\textbf{Llama-4} is a recent family of multimodal models that use interleaved MoE layers within the text encoder. Llama-4 models are distilled from a single larger model with varying number of total parameters and experts. Currently, only the Maverick (400B parameters, 128 experts) and Scout (109B parameters, 16 experts) have been released.\\
\textbf{Mixtral} MoE models were trained in two sizes: 8x7B and 8x22B. Mixtral models are distinct in that they do not use shared experts. They also have the lowest number of total experts (8) among the models in our analysis.

\begin{table}[h!]
\centering
\resizebox{\columnwidth}{!}
{
\begin{tabular}{lccc}
\toprule
\textbf{Model Name} & \makecell{\textbf{Model Total} \\ \textbf{Size (B)}} & \makecell{\textbf{Total} \\ \textbf{Experts}} & \makecell{\textbf{Activated} \\ \textbf{Experts}} \\
\midrule
\textbf{DeepSeek-R1} & \textbf{670} & \textbf{256}  & {8+1}\\
DeepSeek-V2-Lite & 15.7 & 64  & 6+2  \\
Mixtral-8x22B    & 141  & 8   & 2 \\
Mixtral-8x7B     & 46.7 & 8   & 2 \\
Llama-4-Scout    & 109  & 16  & 1+1 \\
Llama-4-Maverick & 400  & 128 & 1+1 \\
\bottomrule
\end{tabular}
}
\caption{Model size and number of experts of the MoE models we study. We denote the number of activated experts for each token as routed + shared.}
\label{tab:model_size}
\end{table}

\subsection{Normalized Overlap Metric}
To account for overlap expected by chance and enable comparison across models with different numbers of total and active experts, we define a chance-corrected overlap score analogous to Cohen’s $\kappa$ and Scott’s $\pi$. 

Let the number of overlapping experts be $o$, the number of active experts per input be $k$, and the total number of experts be $N$. Under a uniform random selection baseline, the expected overlap is:
$\mathbb{E}[o] = \frac{k^2}{N}$.
We define the observed agreement:
$P_o = \frac{o}{k}$
and the expected agreement:
$P_e = \frac{\mathbb{E}[o]}{k} = \frac{k}{N}$.
Then, the normalized overlap score is:

\begin{equation}
\text{score} = \frac{o - \mathbb{E}[o]}{k - \mathbb{E}[o]} = \frac{P_o - P_e}{1 - P_e}
\end{equation}

This is formally equivalent to Cohen’s 
$\kappa = \frac{P_o - P_e}{1 - P_e}
$ and reduces to Scott’s $\pi$ under the assumption of identical marginal distributions. In our setting, $P_e = k/N$ assumes uniform random selection of $k$ experts from a total of $N$ per input. See \S\ref{apd:stats_wic} for a derivation of the random baseline.

\section{Experiment Results and Analysis}
\paragraph{Word-in-Context} 
For 1K pairs of sentences in WiC, we collect router activations for each MoE model (Table~\ref{tab:model_size}) and record the number of overlapping experts at each layer.

We compare the average rate of overlap in sentence pairs where the target word has the same sense versus sentence pairs where it has a different meaning. If sentence pairs where the target word has different senses have higher expert overlap than sentence pairs where the target word has the same sense, then this is evidence that expert routing differentiates on a semantic basis.

\begin{figure}
    \centering
    \includegraphics[width=1.0\linewidth]{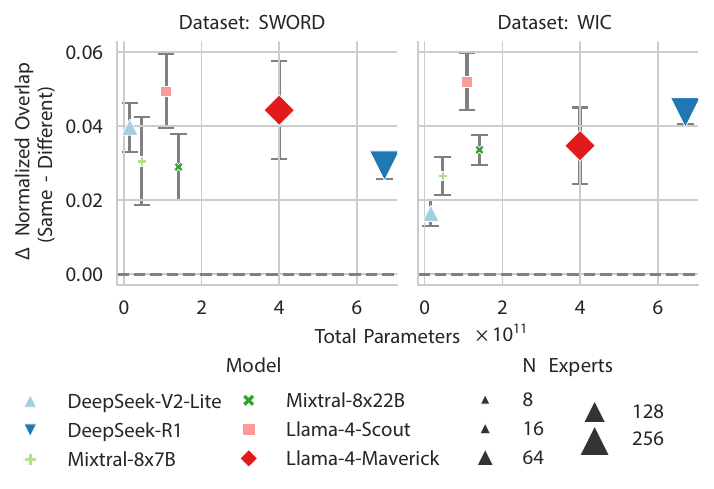}
    \caption{
    The difference between same sense words and different sense words across models and datasets. We find all models show statistically-significantly higher similarity of expert overlap, for same versus differently sensed words, when compared to a baseline of random.}
    \label{fig:treatment_effect}
\end{figure}

\begin{figure}[t]
    \centering
    \includegraphics[width=1.0\linewidth]{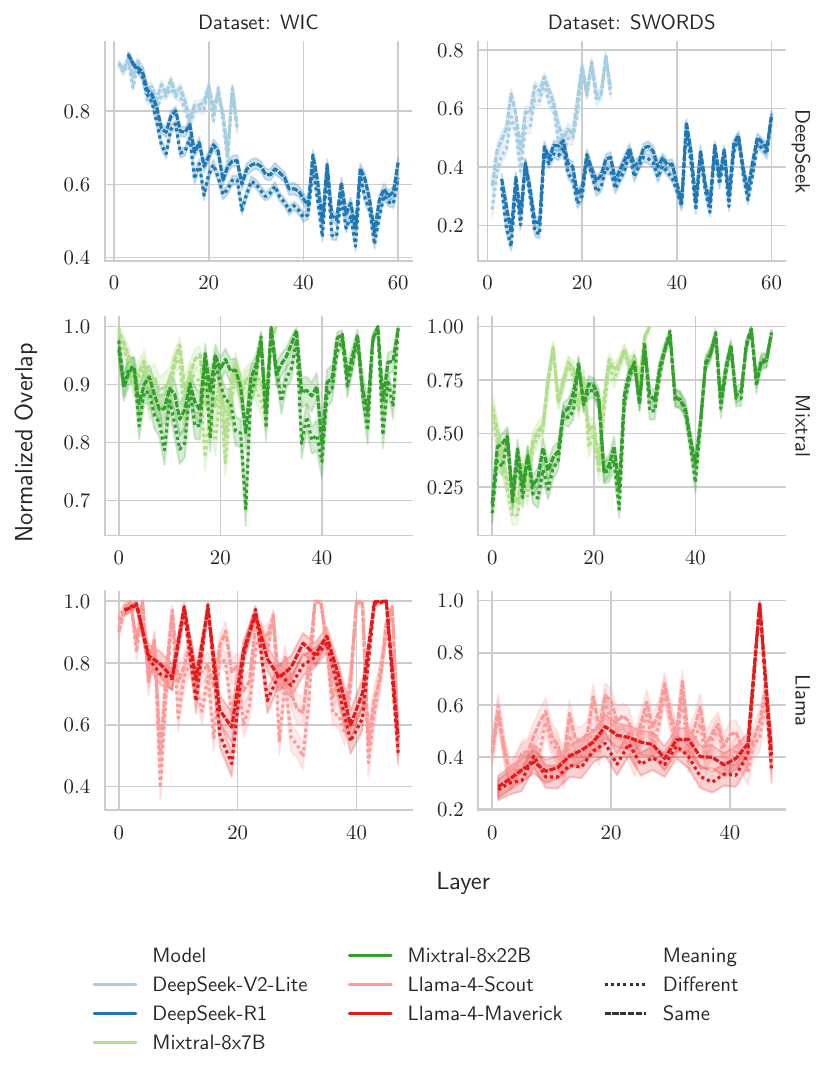} 
    \caption{Layer-wise analysis of MoE LLMs. Generally we find a larger change in overlap for the middle layers (e.g., DeepSeek-R1), and lesser for earlier/later layers. Llama models, with only 1 expert, show much noisier behavior, with an interesting spike in overlap for the penultimate layer.}
    \label{fig:tlayerwise_routing}
\end{figure}

Figure~\ref{fig:tlayerwise_routing} reports, for each layer and model, the mean number of overlapping experts across sentence pairs in the two conditions. We find strong evidence for semantic specialization in these experiments; expert overlap is \textbf{lower} for sentence pairs where the target word has different senses than when they are the same. This effect is statistically significant ($p<0.001$) for all models considered when averaged across all layers. 

For all models the difference in overlap \textit{increases} in intermediary layers. This supports prior findings that semantic features are more salient in the intermediary layers of LLMs ~\citep{niu2022does, kaplan2024tokens}. %
Our results are also suggestive that this pattern emerges at scale; the difference in expert overlap increases with model size. %

\paragraph{SWORDS}

We test whether the equivalent pair has higher expert overlap on average than the lexically different pair for six of our studied models on the test set. We use a paired t-test with the alternative hypothesis that equivalent pair has higher overlap and find strong evidence to reject the null (p $<.0001$) for six all models.

\paragraph{Case Study on Expert overlap in CoT}
We conduct a qualitative analysis using DeepSeek-R1 on DiscoveryWorld \citep{jansen2024discoveryworld}, a large-scale agentic environment suite that tests the abilities of an agent to perform the scientific method. We analyze the degree of expert overlap for different reasoning strategies employed in the CoT. To identify discrete reasoning strategies we analyze the latent representation before routing with a Sparse Autoencoder (SAE) \citep{cunningham2023sparse}. We use the SAE to learn a mapping between the internal activations of R1 and a set of underlying semantic structures. 

By inspecting the trained SAE’s representation during reasoning on the token ``Wait'', we observe that tokens such as ``bet'', ``probably'', and ``attempt'' activate the same SAE feature, suggesting a latent cognitive pattern related to double-checking and uncertainty. This reasoning pattern is most frequently routed to a small subset of experts. We include more examples and details in appendix \S\ref{suppl:sae_disco}.

\section{Conclusion}

Our study provides the first systematic evidence that expert routing in modern Mixture-of-Experts (MoE) language models is sensitive to semantic content.
Across two complementary tasks—word sense disambiguation and lexical substitution—we show that expert overlap increases when meaning is preserved and decreases when it changes.
This effect is robust across six models from three MoE families and persists across model scales and configurations.
We find that semantic routing signals are strongest in the middle layers with
 these signals scaling via model size, suggesting semantic specialization in routing may be a learned, emergent behavior.
Our findings challenge assumptions that routing is primarily token-based and offer a new view on how sparse models organize computation.
By linking routing to semantic similarity, this work enables new directions for interpretability, control, and efficiency in MoE deployment.

\clearpage
\section*{Limitations}
Our analysis is constrained by limited coverage of the MoE design space. Due to the substantial computational cost of training large-scale MoE models, our study relies on a small set of publicly available models, which restricts our ability to assess the effects of broader architectural variations. Additionally, while we focus on architectural differences, variation in training regimes may also influence routing behavior. However, incomplete documentation, particularly regarding optimization strategies such as GRPO, limits our capacity to disentangle these effects or attribute observed patterns to specific training choices.

\section*{Ethics Statement}
For each artifact used e.g. model weights, WiC dataset, and SWORD dataset, we follow the intended use, and while we do not believe that our analysis of these models pose any risks or ethical considerations, we acknowledge the inherent issues with LLMs that are trained on web-scale or biased data. Outputs from LLMs may raise safety concerns due to hallucinations or bias in the training data.

\bibliography{custom}
\bibliographystyle{acl_natbib}

\clearpage
\appendix

\section{Statistical Tests - Random Baseline}
\label{apd:stats_wic}

The baseline number of overlapping experts of we expect to select at random in a given MoE layer can be formalized as follows.
Given independent two draws of $k$ items from $N$ elements (without replacement), the expected number of overlapping items between the two draws can be calculated according to the following formula:

$$
    \mathbb{E}[\text{overlap}] = \frac{k^2}{N}
$$

\begin{proof}
The first draw of $k$ items is at random. For the first item in the second draw, the probability of selecting the same item is $\frac{k}{N}$.

Using the linearity of expectation, the expected total overlap is $\sum_i^k\frac{k}{N}=k\cdot\frac{k}{N}=\frac{k^2}{N}$.
\end{proof}

\begin{figure*}[t]
    \centering
    \includegraphics[width=\linewidth]{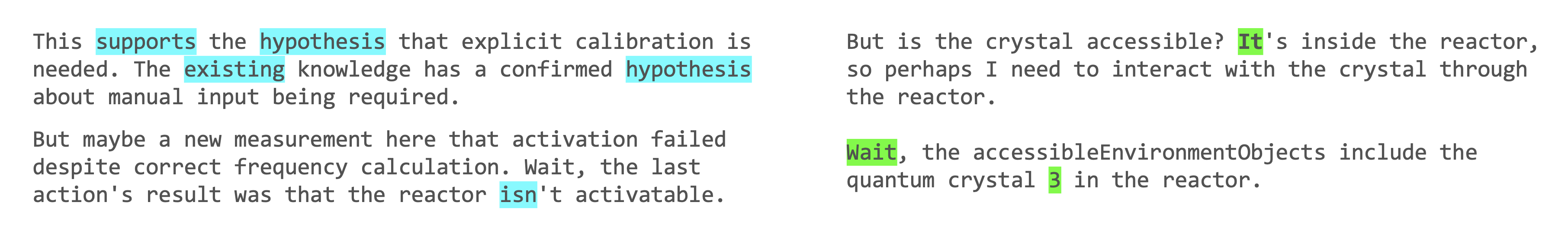} 
    \caption{Left: identified reasoning tokens of SAE head 15376 (highlights indicate non-zero head activation) on DiscoveryWorld chain of thought generations. This head activates when the model analyzes its hypotheses. Right: tokens from SAE head 12649. This head activates when R1 catches an internal reasoning error.}
    \label{fig:two_sae_heads_text_example}
    \vspace{-1em}
\end{figure*}

\section{Additional Qualitative Experiments}
\label{suppl:sae_disco}
DiscoveryWorld \citep{jansen2024discoveryworld}
is a large-scale agentic environment suite that tests the abilities of an agent to perform the scientific method. Each environment has a terminal goal, for example, we study "Reactor Lab" where the agent must tune the frequency of quantum crystals to activate a reactor. To succeed, the agent must formulate and test hypotheses by using available tools, literature, and its own memory. 
Building on the Words-in-Context and SWORDS experiments, we investigate if a similar phenomena of expert specialization can be found for the reasoning patterns that we observe within DeepSeek-R1's CoT. Given any reasoning trace, we find groups of tokens that correspond to a specific reasoning strategy and observe which experts are subsequently activated. If similar experts are used to process all the tokens for a given reasoning strategy, then we have evidence that the experts also specialize by cognitive pattern. 

\subsubsection*{Sparse Autoencoders}
To measure expert overlap, we first need to isolate discrete reasoning patterns to study. To this end, we employ SAEs to learn a mapping between the internal activations of R1 and a set of underlying semantic structures exhibited by the model. 
Briefly, an SAE learns a compressed representation of input vectors $x \in \mathbb{R}^d$. The encoder maps inputs to a higher-dimensional latent space, while the decoder reconstructs the input from the latent representation. Given an encoding dimension $n$, we define the encoder and decoder as:  
$
z = \max(0, W_{\text{enc}} x + b_{\text{enc}})
$
 and
$
\hat{x} = W_{\text{dec}} z
$

where $W_{\text{enc}} \in \mathbb{R}^{n \times d}$ and $W_{\text{dec}} \in \mathbb{R}^{d \times n}$ are the learnable weight matrices of the encoder and decoder respectively, and $b_{\text{enc}} \in \mathbb{R}^{n}$ is a bias term.  
The model is trained using a loss function that balances reconstruction accuracy and sparsity:  
$L = \| x - \hat{x} \|_2^2 + \lambda \| z \|_1$

where the first term is the mean squared error for reconstruction, and the second term is an $L_1$ penalty that encourages sparsity in the latent activations, where we choose $\lambda = 5$ as the trade-off between reconstruction fidelity and sparsity.  

\paragraph{SAE Training}
We evaluate DeepSeek-R1 on the DiscoveryWorld environment: "Reactor Lab", collecting 100 steps through the environment. 
For each step we collect all valid output text including the chain of thought and the corresponding pre-router activations: (the embeddings before expert selection). We consider a generation valid if we have a complete set of "<think>", "</think>" tags. In total we collect 200,000 token-activation pairs. We perform all inference using VLLM~\cite{kwon2023efficient} on Intel\textsuperscript{\textregistered} Gaudi~3 AI accelerators in the Intel\textsuperscript{\textregistered} Tiber\texttrademark~AI Cloud.

We train a standard SAE on these activations using the SAELens library \cite{bloom2024saetrainingcodebase} (MIT License). We trained for 30,000 steps with a batch size of $4096$, learning rate of $5e^{-5}$, SAE width of 28,672, and we reset dead SAE weights after 1K steps. We train the SAE on the activations of layer 7 for a trade-off between early layers with clear token-expert mapping and later layers having high expert selection diversity.

After training, we obtain an atlas that maps individual tokens to higher-level reasoning patterns (see Figure (\ref{fig:two_sae_heads_text_example}) for an example). and show that R1 tends to activate similar experts for all tokens given by single SAE head (neuron), meaning that the experts are not just semantically specialized, but also control the presence of high level reasoning. 

\subsection{DiscoveryWorld Results}
\begin{table}[tb]
    \centering
    \renewcommand{\arraystretch}{.8} %
    \begin{tabular}{ccc}
         Expert 138 & Expert 89 & Expert 81 \\
        \midrule
       reactor    & reactor    & reactor     \\
         core       & microscope & microscope \\
        microscope & ,          & frequency      \\
       it        & it         & maybe      \\
       frequency  & frequency  & crystal       \\
        \bottomrule
    \end{tabular}
    \caption{Top 5 tokens associated with experts often selected for words such as ``hypothesis'' and ``Wait''.} 
    \vspace{-1em}
    \label{tab:expert_top_tokens}
\end{table}

As an illustrative example, we choose two tokens associated with reasoning: ``hypothesis'' and ``Wait''. As a baseline, Table (\ref{tab:expert_top_tokens}) shows an expert-token analysis without an SAE. We see that the experts that are most often allocated for ``Wait'', are also chosen for tokens like ``microscope'', ``frequency'', and ``crystal''. These ancillary tokens are objects/quantities from the environment i.e. the subject of reasoning, but yield no additional information about the reasoning process itself. %

\begin{table}[]
    \centering
    \begin{tabular}{lcc}
        \makecell{Input \\ Token} & \makecell{SAE \\ Value}  & Top 5 occurring experts \\
        \toprule
        bet      & 17.16  & 47 \hfill \textbf{133} \hfill \textbf{136} \hfill \textbf{138} \hfill 148\\
         Wait     & 7.94   & 81 \hfill \textbf{89} \hfill 95 \hfill \textbf{133} \hfill \textbf{136}\\
         notes    & 6.79   & 71 \hfill \textbf{89} \hfill 90 \hfill \textbf{133} \hfill \textbf{138}\\
         probably & 4.97   & 48 \hfill 57 \hfill 101 \hfill \textbf{136} \hfill \textbf{138}\\
         output   & 4.59   & 81 \hfill \textbf{89} \hfill \textbf{133} \hfill \textbf{136} \hfill \textbf{138}\\
         3        & 3.92   & 81 \hfill \textbf{89} \hfill 95 \hfill \textbf{136} \hfill \textbf{138}\\
         fail     & 3.53   & 81 \hfill \textbf{89} \hfill 121 \hfill \textbf{133} \hfill \textbf{136}\\
         It       & 2.87   & \textbf{89} \hfill \textbf{133} \hfill \textbf{136} \hfill \textbf{138} \hfill 183\\
         ones     & 2.06   & 57 \hfill 101 \hfill 121 \hfill \textbf{133} \hfill \textbf{136}\\
         attempt  & 1.72   & 15 \hfill 81 \hfill \textbf{89} \hfill 95 \hfill \textbf{133}\\

        \bottomrule
    \end{tabular}
    \caption{We selected the top activating SAE head on the word "Wait" and used its activations to identify additional activating tokens. We find the top 5 occurring experts given these tokens is highly consistent, experts chosen for 50\% or more tokens are bolded.}
    \vspace{-1em}

    \label{tab:sae_Wait_main}
\end{table}

The SAE provides further insight by examining sets of tokens that are linked through the maximal activation of a single SAE head. 
Table (\ref{tab:sae_Wait_main}) shows an example where a single head (active on “Wait”) identifies semantically similar tokens. By inspecting the corresponding SAE activations, we observe tokens such as “bet,” “probably,” and “attempt,” which suggest a cognitive pattern of uncertainty regarding the current strategy. Moreover, we find that this reasoning pattern is most commonly routed to a small set of experts. Examining these tokens and activations in context (e.g., see Figure (\ref{fig:two_sae_heads_text_example})) further illustrates how R1 leverages contextual information in its reasoning process.

We also find that the SAE head corresponding to ``hypothesis'', yields a pattern of overlapping experts along semantically similar tokens such as: ``definitely'', ``perform'', ``analyzing'', ``scientific'', and ``information''. In summation, we find that R1 consistently chooses a small set of experts for reasoning patterns identified by the SAE, indicating that the experts also specialize by thought process.

\subsection{SAE token analysis} 
\label{apd:more_sae_tables}
\begin{table}[]
    \centering
    \begin{tabular}{lcc}
        \makecell{Input \\ Token} & \makecell{SAE \\ Value}  & Top 5 occurring experts \\
        \toprule
         wait & 14.97 & 47 \hfill 133 \hfill 138 \hfill 148 \hfill 183 \\
         Are & 1.7 & 90 \hfill 133 \hfill 136 \hfill 138 \hfill 170 \\
         ones & 1.24 & 57 \hfill 101 \hfill 121 \hfill 133 \hfill 136 \\
         No & 0.32 & 26 \hfill 47 \hfill 136 \hfill 138 \hfill 183 \\
         best & 0.16 & 15 \hfill 47 \hfill 81 \hfill 89 \hfill 133 \\
         attempt & 0.05 & 15 \hfill 81 \hfill 89 \hfill 95 \hfill 133 \\
         Wait & 0.02 & 81 \hfill 89 \hfill 95 \hfill 133 \hfill 136 \\
        \bottomrule
    \end{tabular}
    \caption{An analysis of selected experts by leveraging the trained Sparse Autoencoder. The target token is "wait."}
    \label{tab:sae_wait}
\end{table}

\begin{table}[]
    \centering
    \begin{tabular}{lcc}
        \makecell{Input \\ Token} & \makecell{SAE \\ Value}  & Top 5 occurring experts \\
        \toprule
         giving & 4.47 & 11 \hfill 15 \hfill 81 \hfill 89 \hfill 90 \\
         hypothesis & 4.04 & 11 \hfill 15 \hfill 81 \hfill 89 \hfill 90 \\
         definitely & 2.26 & 11 \hfill 15 \hfill 81 \hfill 89 \hfill 90 \\
         perform & 1.96 & 11 \hfill 15 \hfill 81 \hfill 89 \hfill 90 \\
         priority & 1.82 & 11 \hfill 15 \hfill 81 \hfill 89 \hfill 90 \\
         analyzing & 1.51 & 11 \hfill 15 \hfill 81 \hfill 89 \hfill 90 \\
         scientific & 1.17 & 11 \hfill 15 \hfill 81 \hfill 89 \hfill 90 \\
        \bottomrule
    \end{tabular}
    \caption{An analysis of selected experts by leveraging the trained Sparse Autoencoder. The target token is "hypothesis."}
    \label{tab:sae_hypothesis}
\end{table}

\begin{table}[]
    \centering
    \begin{tabular}{lcc}
        \makecell{Input \\ Token} & \makecell{SAE \\ Value}  & Top 5 occurring experts \\
        \toprule
        combining & 13.50 & 11 \hfill 15 \hfill 69 \hfill 90 \hfill 136\\
        formatted & 13.32 & 11 \hfill 15 \hfill 69 \hfill 90 \hfill 136\\
        frequencies & 13.31 & 11 \hfill 15 \hfill 69 \hfill 90 \hfill 136\\
        accessible & 13.31 & 11 \hfill 15 \hfill 26 \hfill 136 \hfill 138\\
        restrictions & 13.29 & 11 \hfill 15 \hfill 26 \hfill 136 \hfill 138\\
        rejected & 13.13 & 11 \hfill 15 \hfill 69 \hfill 90 \hfill 136\\
        559 & 9.92 & 11 \hfill 15 \hfill 69 \hfill 90 \hfill 136\\
        UUID & 6.83 & 11 \hfill 15 \hfill 26 \hfill 136 \hfill 138\\
        854 & 6.62 & 11 \hfill 15 \hfill 69 \hfill 90 \hfill 136\\
        obtaining & 6.44 & 15 \hfill 90 \hfill 95 \hfill 136 \hfill 138\\
        \bottomrule
    \end{tabular}
    \caption{An analysis of selected experts by leveraging the trained Sparse Autoencoder. We selected the top activating SAE head on the word "UUID" and used its activation's value to identify other semantically similar tokens. The top 5 occurring experts are highly consistent across these varying words.}
    \label{tab:sae_uuid}
\end{table}

In tables~(\ref{tab:sae_wait}, \ref{tab:sae_hypothesis}, \ref{tab:sae_uuid})
we show top experts by leveraging SAE activations on a selection of hand chosen interesting tokens. We find striking consistency across expert selection when using the SAE to find semantically similar concepts.

\section{DiscoveryWorld Environment Details}
\label{app:discodetails}
DiscoveryWorld features 8 tasks centered on different scientific fields. We choose to evaluate R1 on the "Reactor Lab" environment, where the stated goal is to: ``discover a relationship (linear or quadratic) between a physical crystal property (like temperature or density) and its resonance frequency through regression, and use this to tune and activate a reactor.''

\begin{figure}[ht]
    \centering
    \includegraphics[width=\linewidth]{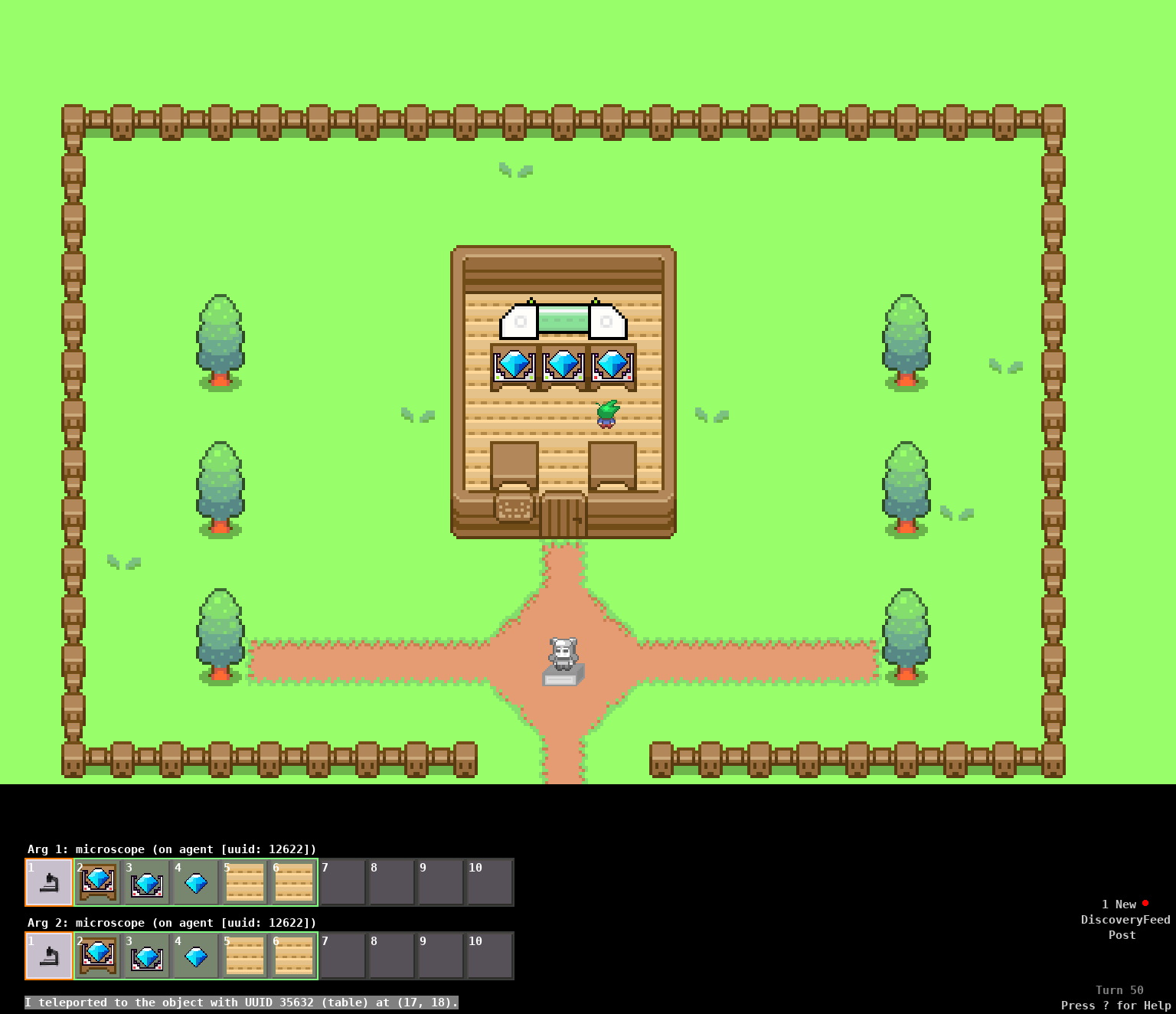} 
    \caption{Visual observation in the Reactor Lab environment at step 50.}
    \label{fig:dw_reactor_pic}
\end{figure}

In Figure (\ref{fig:dw_reactor_pic}), we show the Reactor Lab environment, where the agent has access the crystals and microscope in its inventory. The pixel-based visual observation itself it not used by R1 directly, but the prompt (see below) contains a structured description of the environment. 

We show an example prompt and chain of thought output by R1 in the Reactor Lab environment below. 

\onecolumn

\begin{tcolorbox}[colback=green!5!white, colframe=green!75!black, title=Example Prompt on DiscoveryWorld Reactor Lab, breakable]
{\small  \verbatiminput{prompts.txt}}
\end{tcolorbox}

\begin{tcolorbox}[colback=orange!5!white, colframe=orange!75!black, title=Example Reasoning Output from DeepSeek-R1 (step 50), breakable, width=\textwidth]
{\small  \verbatiminput{cot.txt}}
\end{tcolorbox}

\end{document}